\documentclass[11pt,letterpaper]{article}
\usepackage{emnlp2016}
\usepackage{times}
\usepackage{url}
\usepackage{latexsym}
\usepackage{setspace}
\usepackage{amsmath}
\usepackage{amssymb}
\usepackage{stmaryrd}
\usepackage{array}
\usepackage{graphicx}
\usepackage{cases}
\usepackage{watermark}
\usepackage{pbox}
\usepackage{multirow}
\usepackage{soul}
\setulcolor{red}
\setstcolor{red}
\setul{}{.3ex}
\usepackage{enumitem}
\usepackage{xspace}
\usepackage{makecell}
\usepackage{array}
\usepackage{diagbox}
\makeatletter
\newcommand{\thickhline}{%
    \noalign {\ifnum 0=`}\fi \hrule height 1pt
    \futurelet \reserved@a \@xhline
}
\newcolumntype{"}{@{\hskip\tabcolsep\vrule width 1pt\hskip\tabcolsep}}
\makeatother

\xspaceaddexceptions{$\lambda$} 

\newcommand{\squeeze}[1]{\hspace{-1mm}#1\hspace{-1mm}}

\newcommand{\msq}{\text{M}\ensuremath{^2}\xspace}

\newcommand{\heilman}{LFM\xspace}

\newcommand{\interpolation}{S_{\text{I}}}
\newcommand{\referencemetric}{S_{\text{R}}} 
\newcommand{\featuremetric}{S_{\text{G}}} 
\newcommand{\erater}{ER\xspace}
\newcommand{\lt}{LT\xspace}
\newcommand{\eratername}{e-rater\textsuperscript{\textregistered}\xspace}

\usepackage{xcolor}

\makeatletter
\newcommand{\@BIBLABEL}{\@emptybiblabel}
\newcommand{\@emptybiblabel}[1]{}
\makeatother
\usepackage[hidelinks]{hyperref}

\emnlpfinalcopy

\def\emnlppaperid{677}

\title{There's No Comparison: Reference-less Evaluation Metrics\\in Grammatical Error Correction}

\newcommand{\clsp}{\ensuremath{{}^\text{1}}}
\newcommand{\grammarly}{\ensuremath{{}^\text{2}}}

\author{
  Courtney Napoles\textnormal{,\clsp}
   Keisuke Sakaguchi\textnormal{,\clsp} \and
   Joel Tetreault\textnormal{\grammarly}\\
 \clsp Center for Language and Speech Processing, Johns Hopkins University \\
 \grammarly Grammarly\\
 \tt{\{napoles,keisuke\}@cs.jhu.edu},
 \url{joel.tetreault@grammarly.com}}

\date{}

\begin{document}
\maketitle

\begin{abstract}

Current methods for automatically evaluating grammatical error correction
(GEC) systems rely on gold-standard references.  However,
these methods suffer from penalizing grammatical
edits that are correct but not in the gold standard.  We show
that reference-less grammaticality metrics correlate very strongly with human judgments 
and are competitive with the leading reference-based evaluation metrics.
By interpolating both methods, we achieve state-of-the-art correlation with human judgments.
Finally, we show that GEC metrics are much more reliable when they are calculated at the sentence level instead of the corpus level. 
We have set up a CodaLab site for benchmarking GEC output using a common dataset and different evaluation metrics.

\end{abstract}




\section{Introduction}
\label{sec:intro}

Grammatical error correction (GEC) has been evaluated by comparing the changes made by a system to the corrections made in gold-standard annotations.
Following the recent shared tasks in this field (e.g., Ng et al. \shortcite{ng-EtAl:2014:W14-17}), several papers have critiqued GEC metrics and proposed new methods.  
Existing metrics depend on gold-standard corrections and therefore have a notable weakness: systems are penalized for making corrections that do not appear in the references.\footnote{We refer to the gold-standard corrections as \textit{references} because \textit{gold standard} suggests just one accurate correction.}
For example, the following output has low metric scores even though three appropriate corrections were made to the input:

\noindent
\begin{small}
\begin{tabular}{|p{7.3cm}|}
    \hline
    However , people now can \st{contact} \textbf{{\color{red}communicate}} with \st{anyone} \textbf{{\color{red}people}} all over the world who can use computer\textbf{{\color{red}s}} at any time , and there is no time delay of messages .\\
    \hline
\end{tabular} 
\end{small}
\vspace{1mm}

\noindent
These changes (in red) were not seen in the references and therefore the 
metrics GLEU and \msq (described in \S \ref{sec:prior}) score this output worse than 75\% of 15,000 other generated sentences.




While grammaticality-based, reference-less metrics have been effective in estimating the quality
of machine translation (MT) output, the utility of such metrics has not been 
investigated previously for GEC.  We hypothesize that 
such methods can overcome this weakness in  reference-based GEC
metrics.  This paper has four contributions:  1) We develop three {\em grammaticality} metrics that are competitive with current reference-based measures and correlate very strongly with human judgments. 2) We achieve state-of-the-art performance when interpolating a leading reference-based metric with a grammaticality metric. 3) We identify an interesting result that the mean of sentence-level scores is substantially better for evaluating systems than the system-level score.  4) We release code for two grammaticality metrics and establish an online platform for evaluating GEC output.


\section{Prior work}
\label{sec:prior}

To our knowledge, this is the first work to evaluate GEC without references.
Within MT, this task is called \textit{quality estimation}.
MT output is evaluated by its \textit{fluency}, or adherence to accepted conventions of grammaticality and style, and \textit{adequacy}, which is the input's meaning conveyed in the output.  
Quality estimation targets fluency by estimating the amount of post-editing needed by the output.
This has been the topic of recent shared tasks, e.g. \newcite{bojar-EtAl:2015:WMT}.
\nocite{callisonburch-EtAl:2012:WMT,bojar-EtAl:2013:WMT,bojar-EtAl:2014:W14-33}
\newcite{specia2010machine} evaluated the quality of translations using sentence-level features from the output but not the references, predicting discrete 
and continuous 
scores.
A strong baseline, \textit{QuEst}, uses support vector regression trained over 17 features extracted from the output \cite{specia-EtAl:2013:SystemDemo}. 
Most closely related to our work, \newcite{parton2011rating} applied features from Educational Testing Service's \eratername \cite{attali2006automated}
to evaluate MT output with a ranking SVM, without references, and improved performance by
 including features derived from MT metrics (BLEU, TERp, and METEOR). 

Within the GEC field, recent shared tasks  
have prompted the development and scrutiny of new metrics for evaluating GEC systems.
The Helping Our Own shared tasks evaluated systems using precision, recall, and F-score against annotated gold-standard corrections \cite{dale-kilgarriff:2011:ENLG,dale-anisimoff-narroway:2012:BEA}.
The subsequent CoNLL Shared Tasks on GEC \cite{ng-EtAl:2013:CoNLLST,ng-EtAl:2014:W14-17} were scored with the MaxMatch metric (\msq), which captures word- and phrase-level edits by calculating the F-score over an edit lattice \cite{dahlmeier-ng:2012:NAACL-HLT}.
\newcite{felice-briscoe:2015:NAACL} identified shortcomings of \msq and proposed I-measure to address these issues. 
I-measure computes the accuracy of a token-level alignment between the original, generated, and gold-standard sentences.
These precision- and recall-based metrics measure fluency and adequacy by penalizing inappropriate changes, which alter meaning or introduce other errors. 
Changes consistent with the annotations indicate improved fluency and no change in meaning.

Unlike these metrics, GLEU scores output by penalizing n-grams found in the input and output but not the reference \cite{napoles-EtAl:2015:ACL-IJCNLP}.
Like BLEU \cite{papineni-EtAl:2002:ACL}, GLEU captures both fluency and adequacy with n-gram overlap.
Recent work has shown that GLEU has the strongest correlation with human judgments compared to the GEC metrics described above \cite{tacl-gec-eval-2016}.
These GEC metrics are all defined at the corpus level, meaning that the statistics are accumulated over the entire output and then used to calculate a single system score.


\section{Explicitly evaluating grammaticality}
\label{sec:gbm}

GLEU, I-measure, and \msq are calculated based on comparison to reference corrections. These Reference-Based Metrics (\textit{RBMs}) credit corrections seen in the references and penalize systems for ignoring errors and making bad changes (changing a span of text in an ungrammatical way or introducing errors to grammatical text).
However, RBMs make two strong assumptions: that the annotations in the references are \textit{correct} and that they are \textit{complete}. 
We argue that these assumptions are invalid and point to a deficit in current evaluation practices.
In GEC, the agreement between raters can be low due to the challenging nature of the task \cite{bryant-ng:2015:ACL-IJCNLP,rozovskaya-roth:2010:BEA,tetreault-chodorow:2008:HJCL},  indicating that annotations may not be correct or complete. 

An exhaustive list of all possible corrections would be time-consuming, if not impossible.
As a result, RBMs penalize output that has a valid correction that is not present in the references or that addresses an error not corrected in the references.
The example in \S \ref{sec:intro} has low GLEU and \msq scores, even though the output addresses two errors (GLEU=0.43 and \msq = 0.00, in the bottom half and quartile of 15k system outputs, respectively).

To address these concerns, we propose three metrics to evaluate the grammaticality of output without comparing to the input or a gold-standard sentence (Grammaticality-Based Metrics, or \textit{GBMs}).
We expect GBMs to score sentences, such as our example in \S 1, more highly.
The first two metrics are scored by counting the errors found by existing grammatical error detection tools.
The error count score is simply calculated:  $ 1 - \frac{\textrm{\# errors}}{\textrm{\# tokens}}$.
Two different tools are used to count errors: \eratername's grammatical error detection modules (\erater) and Language Tool \cite{milkowski2010developing} (\lt).
We choose these because, while \eratername is a large-scale, robust tool that detects more errors than Language Tool,\footnote{In the data used for this work, \eratername detects approximately 15 times more errors than Language Tool.} 
it is proprietary whereas Language Tool is publicly available and open sourced.

For our third method, we estimate a grammaticality score with a linguistic feature-based model (\heilman), which is our implementation of \newcite{heilman-EtAl:2014:P14-2}.\footnote{Our implementation is slightly modified in that it does not use features from the PET HPSG parser.}  
The \heilman score is a ridge regression over a variety of linguistic features related to grammaticality, including the number of misspellings, language model scores, OOV counts, and PCFG and link grammar features.
It has been shown to effectively assess the grammaticality of learner writing.
LFM predicts a score for each sentence while ER and LT, like the RBMs, can be calculated with either sentence- or document-level statistics.
To be consistent with LFM, for all metrics in this work we score  each sentence individually and report the system score as the mean of the sentence scores.  
We discuss the effects of modifying metrics from a corpus-level to a sentence-level in \S 5.


Consistent with our hypothesis, \erater and \lt score the \S 1 example in the top quartile of outputs and \heilman ranks it in the top half. 
\subsection{A hybrid metric}
\label{sect:interpol}

The obvious deficit of GBMs is that they do not measure the adequacy of generated sentences, so they could easily be manipulated with grammatical output that is unrelated to the input.
An ideal GEC metric would measure both the grammaticality of a generated sentence and its meaning compared to the original sentence, and would not necessarily need references.
The available data of scored system outputs are insufficient for developing a new metric due to their limited size, thus we turn to interpolation to develop a sophisticated metric that jointly captures grammaticality and adequacy. 

To harness the advantage of RBMs (adequacy) and GBMs (fluency), we build combined metrics, interpolating each RBM with each GBM. 
For a sentence of system output, the interpolated score ($\interpolation$) of the GBM score ($\featuremetric$) and RBM score ($\referencemetric$) is computed as follows: 
\vspace{-2mm}
\begin{equation*}
\interpolation = (1-\lambda)\featuremetric+\lambda\referencemetric
\end{equation*}
All values of $\featuremetric$ and $\referencemetric$ are in the interval $[0, 1]$, except for I-measure, which falls between $[-1, 1]$, and the distribution varies for each metric.\footnote{Mean scores are GLEU $0.52\pm 0.21$, \msq\ $0.21\pm 0.34$, IM $0.10\pm 0.30$, ER $0.91\pm 0.10$, LFM $0.50\pm 0.16$, LT $1.00\pm 0.01$.}
The system score is the average $\interpolation$ of all generated sentences.
\section{Experiments}
\label{sec:experiments}

To assess the proposed metrics, we apply the RBMs, GBMs, and interpolated metrics to score the output of 12 systems participating in the CoNLL-2014 Shared Task on GEC \cite{ng-EtAl:2014:W14-17}.  
Recent works have evaluated RBMs by collecting human rankings of these system outputs and comparing them to the metric rankings \cite{grundkiewicz-junczysdowmunt-gillian:2015:EMNLP,napoles-EtAl:2015:ACL-IJCNLP}.
In this section, we compare each 
metric's ranking to the 
human ranking of \newcite[Table 3c]{grundkiewicz-junczysdowmunt-gillian:2015:EMNLP}.
We use 20 references for scoring with RBMs: 2 original references, 10 references collected by \newcite{bryant-ng:2015:ACL-IJCNLP}, and 8 references collected by \newcite{tacl-gec-eval-2016}. 
The motivations for using 20 references are twofold: the best GEC evaluation method uses these 20 references with the GLEU metric \cite{tacl-gec-eval-2016}, and work in machine translation shows that more references are better for evaluation \cite{FinchAS04}. 
Due to the low agreement discussed in \S \ref{sec:gbm}, having more references can be beneficial for evaluating a system when there are multiple viable ways of correcting a sentence.  
Unlike previous GEC evaluations, all metrics reported here use the \textit{mean} of the sentence-level scores for each system. 




\begin{table}[t]
	\footnotesize
    \centering
    \begin{tabular}{l|c|c}
        \hline
        \textbf{Metric} & \textbf{Spearman's} $\rho$ & \textbf{Pearson's} $r$   \\
        \hline
        GLEU            & \textbf{0.852} & \textbf{0.838} \\ 
        \erater         & \textbf{0.852} & 0.829 \\
        \lt             & 0.808 & 0.811 \\
        I-measure       & 0.769 & 0.753 \\ 
        \heilman        & 0.780 & 0.742 \\
        \msq            & 0.648 & 0.641 \\ 
        \hline        
    \end{tabular}
    \caption{\label{tab:noref-corr} Correlation between the human and metric rankings.}
\end{table}

\begin{table*}[th!]
    \centering
    \footnotesize
    \begin{tabular}{p{.7cm}|| l " l l l || l " l l l}
        \hline
    & \multicolumn{4}{c||}{\textit{ranked by Spearman's rank coefficient ($\rho$)}} & \multicolumn{4}{c}{\textit{ranked by Pearson's correlation coefficient ($r$)}}\\
    \thickhline
		        &            			& \textbf{ER}           	& \textbf{LFM}  & \textbf{LT}      &            			& \textbf{ER}           	& \textbf{LFM}   & \textbf{LT} \\
    \hline
          		& \textit{no intrpl.} 	& 0.852 (0)		& 0.780 (0)    	& 0.808 (0)    	& \textit{no intrpl.} 	& 0.829 (0)    		& 0.742 (0)    	& 0.811 (0) \\ \thickhline
    \textbf{GLEU}  	& 0.852 (1) 	& {\bf 0.885} (0.03) 	& 0.874 (0.27) 	& 0.857 (0.04)	& 0.838 (1) 		& {\bf 0.867} (0.27) 	& 0.845 (0.84)	& 0.867 (0.09)       \\
    \textbf{I-m.}      	& 0.769 (1)  	& 0.874 (0.19) 		& 0.863 (0.37) 	& 0.852 (0.01) 	& 0.753 (1)  		& 0.837 (0.02) 		& 0.791 (0.22) 	& 0.828 (0.01)      \\
    \rule{0pt}{2.2ex}\textbf{\msq} 
    				& 0.648 (1)  	& 0.868 (0.01) 		& 0.852 (0.05) 	& 0.808 (0.00) 	& 0.641 (1)  		& 0.829 (0.00) 		& 0.754 (0.04) 	& 0.811 (0.00)  \\ \thickhline
    \end{tabular}
    \caption{\label{tab:interpolation} Oracle correlations between interpolated metrics and the human rankings. The $\lambda$ value for each metric is in parentheses.} 
\end{table*}

Results are presented in Table \ref{tab:noref-corr}.
The error-count metrics, \erater and \lt, have stronger correlation than all RBMs except for GLEU, which is the current state of the art.  
GLEU has the strongest correlation with the human ranking ($\rho=0.852$, $r=0.838$), followed closely by \erater, which has slightly lower Pearson correlation ($r=0.829$) but equally as strong rank correlation, which is arguably more important when comparing different systems.
I-measure and LFM have similar strength correlations, and \msq is the lowest performing metric, even though its correlation is still strong ($\rho=0.648$, $r=0.641$).

\begin{table}[t]
\centering
\footnotesize
\begin{tabular}{c|c|p{5.3cm}}
\hline
\squeeze{\textbf{GLEU}}	& \squeeze{\textbf{Intrpl.}}	& \\
\textbf{rank}	& \textbf{rank}	& \textbf{Output sentence} \\
\hline
1	& 2   & \textit{Genectic} testing is a personal decision , with the knowledge that there is a \textit{possiblity} that one could be a carrier or not .  \\ \hline
2	& 3      &    \textit{Genectic} testing is a personal decision , the \textit{kowledge} that there is a \textit{possiblity} that one could be a carrier or not .   \\\hline
3	& 1   & Genetic testing is a personal decision , with the knowledge that there is a possibility that one could be a carrier or not .  \\\hline
\end{tabular}
\renewcommand{\baselinestretch}{0.8}
\caption{\label{tab:example} An example of system outputs ranked by GLEU and GLEU interpolated with ER. Words in italics are misspelled.}
\end{table}

Next we compare the interpolated scores with the human ranking,
testing 101 different values of $\lambda$ [0,1] to find the oracle value.  
Table \ref{tab:interpolation} shows the correlations between the human judgments and the oracle interpolated metrics.
Correlations of interpolated metrics are the upper bound and the correlations of uninterpolated metrics (in the first column and first row) are the lower bound.
Interpolating GLEU and IM with GBMs has stronger correlation than any uninterpolated metric (i.e. $\lambda=0$ or 1),  and the oracle interpolation of \erater and GLEU manifests the strongest correlation with the human ranking ($\rho=0.885$, $r = 0.867$).\footnote{To verify that these metrics cannot be gamed, we interpolated GBMs with RBMs scored against randomized references, and got scores 15\% lower than un-gamed scores, on average.}
\msq has the weakest correlation of all uninterpolated metrics and, when combined with GBMs, three of the interpolated metrics have $\lambda = 0$, meaning the interpolated score is equivalent to the GBM and \msq does not contribute.

Table \ref{tab:example} presents an example of how interpolation can help evaluation.
The top two sentences ranked by GLEU have misspellings that were not corrected in the NUCLE references.
Interpolating with a GBM rightly ranks the misspelled output below the corrected output.

\begin{figure}
\vspace{-6mm}
\hspace*{2mm}\includegraphics[width=80mm]{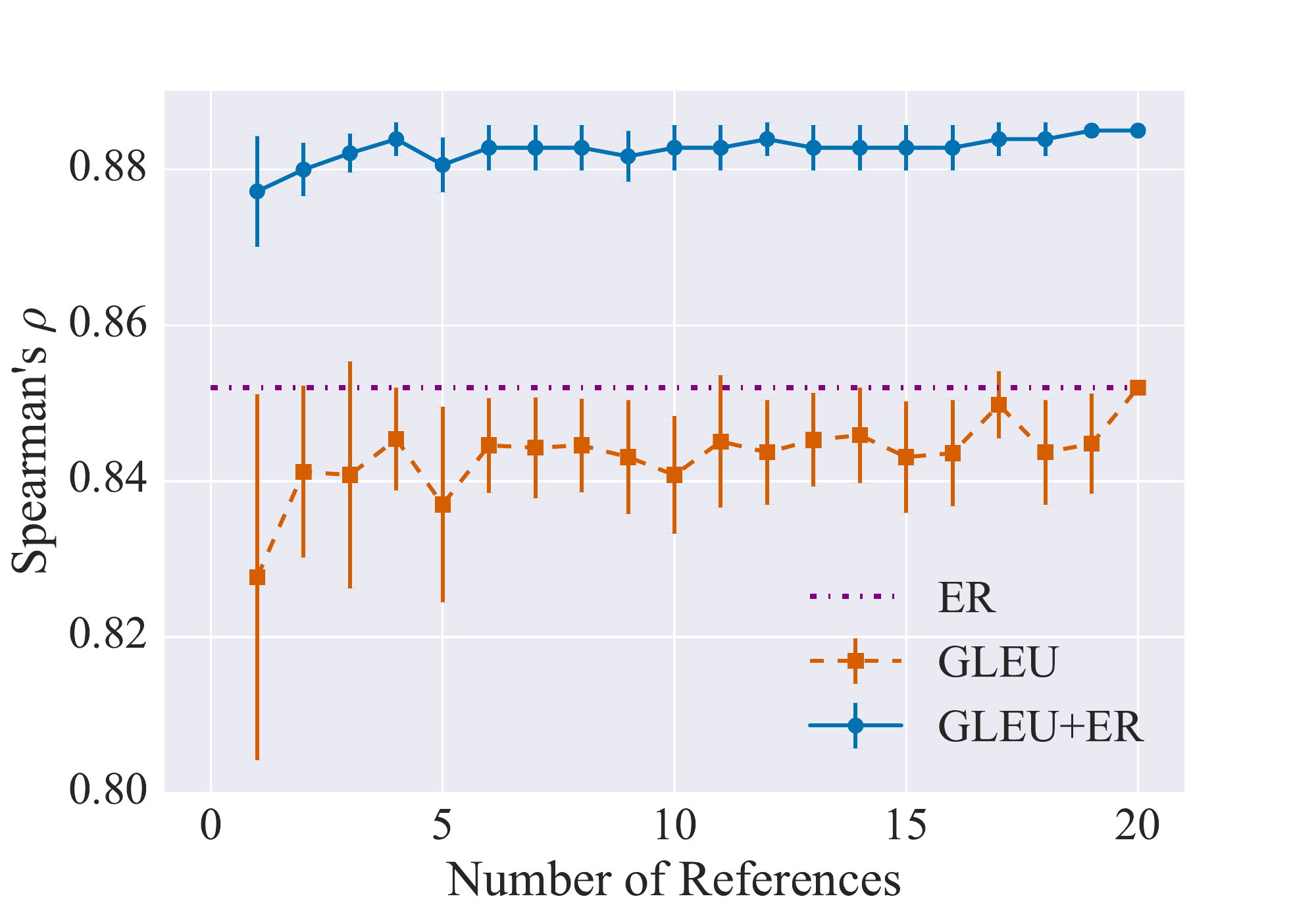}
\vspace{-5mm}
 \renewcommand{\baselinestretch}{0.8}
\caption{\label{fig:num-refs} The mean correlation of oracle interpolated GLEU and ER scores across different reference sizes, compared to the uninterpolated metrics. Bars indicate a 95\% confidence interval.}
\vspace{-1.2mm}
\end{figure}

Since these experiments use a large number of references (20), we determine how different reference sizes affect the interpolated metrics 
by systematically increasing the number of references from 1 to 20 and randomly choosing $n$ references to use as a gold standard when calculating the RBM scores, repeating 10 times for each value $n$ (Figure \ref{fig:num-refs}).
The correlation is nearly as strong with 3 and 20 references ($\rho=0.884$ v. 0.885), and interpolating with just 1 reference is nearly as good (0.878) and improves over any uninterpolated metric.

We acknowledge that using GBMs is in effect using GEC systems to score other GEC systems. Interestingly, we find that even if the GBMs are imperfect, they still boost performance of the RBMs. GBMs have been trained to recognize errors in different contexts and, conversely, can identify correct grammatical constructions in diverse contexts, where the RBMs only recognize corrections made that appear in the gold standards, which are limited.  Therefore GBMs can make a nice complement to  shortcomings that RBMs may have.

\section{Sentence-level evaluation}


In the course of our experiments, we noticed that \mbox{I-measure} and GLEU have stronger correlations with the expert human ranking when using the mean sentence-level score (Table \ref{tab:corpus-scores}).\footnote{The correlations in Table \ref{tab:corpus-scores} differ from what was reported in \newcite{grundkiewicz-junczysdowmunt-gillian:2015:EMNLP} and \newcite{napoles-EtAl:2015:ACL-IJCNLP} due to the references and sentence-level computation used in this work. }
Most notably, I-measure does not correlate at all as a corpus-level metric but has a very strong correlation at the sentence-level (specifically, $\rho$ improves from -0.055 to 0.769).
This could be because corpus-level statistics equally distribute counts of correct annotations over all sentences, even those where the output neglects to make any necessary corrections. 
Sentence-level statistics would not average the correct counts over all sentences in this same way.
As a result, a corpus-level statistic may over-estimate the quality of system output. 
Deeper investigation into this phenomenon is needed to understand why the mean sentence-level scores do better.


\begin{table}
\footnotesize
\centering
\begin{tabular}{l"c|c"c|c}
\hline
& \multicolumn{2}{c"}{\textbf{Corpus}}	& \multicolumn{2}{c}{\textbf{Sentence}}\\

\hline
\textbf{Metric}	& $\rho$    & $r$   & $\rho$ & $r$   \\ 
\thickhline
GLEU			& 0.725		& 0.724 & 0.852 & 0.838 \\
I-m.			& -0.055$^*$& 0.061 & 0.769 & 0.753 \\
\msq			& 0.692		& 0.617 & 0.648 & 0.641 \\
\hline
\end{tabular}
  \renewcommand{\baselinestretch}{0.8}
\caption{\label{tab:corpus-scores} Correlation with human ranking when using corpus- and sentence-level metrics. $^*$ indicates a significant difference from the corresponding sentence-level correlation  ($p<0.05$).\protect\footnotemark[7]}
\end{table}
\footnotetext[7]{Significance is found by applying a two-tailed t-test to the Z-scores attained using Fisher's z-transformation.}

\section{Summary}

We have identified a shortcoming of reference-based metrics: they penalize changes made that do not appear in the references, even if those changes are acceptable.
To address this problem, we developed metrics to explicitly measure grammaticality without relying on reference corrections and showed that the error-count metrics are competitive with the best reference-based metric.
Furthermore, by interpolating RBMs with GBMs, the system ranking has even stronger correlation with the human ranking.
The ER metric, which was derived from counts of errors detected using \eratername, is nearly as good as the state-of-the-art RBM (GLEU) and the interpolation of these metrics has the strongest reported correlation with the human ranking ($\rho=0.885$, $r=0.867$). 
However, since \eratername is not widely available to researchers, we also tested a metric using Language Tool, which does nearly as well when interpolated with GLEU ($\rho=0.857$, $r=0.867$).

Two important points should be noted: First, due to the small sample size (12 system outputs), none of the rankings significantly differ from one another except for the corpus-level I-measure.
Secondly, GLEU and the other RBMs already have strong to very strong correlation with the human judgments, which makes it difficult for any combination of metrics to perform substantially higher.  
The best uninterpolated and interpolated metrics use an extremely large number of references (20), however Figure \ref{fig:num-refs} shows that interpolating GLEU using just one reference has stronger correlation than any uninterpolated metric.  This supports the use of interpolation to improve GEC evaluation in any setting.


This work is the first exploration into applying fluency-based metrics to GEC evaluation. 
We believe that, for future work, fluency measures could be further improved with other methods, such as using existing GEC systems to {\em detect} errors, or even using an ensemble of systems to improve coverage (indeed, ensembles have been useful in the GEC task itself \cite{susanto-phandi-ng:2014:EMNLP2014}).  There is also recent work from the MT community, such as the use of 
confidence bounds \cite{graham-liu:2016:N16-1} or uncertainty measurement \cite{beck-specia-cohn:2016:CoNLL},
which could be adopted by the GEC community.


Finally, in the course of our experiments, we determined that metrics calculated on the sentence-level is more reliable for evaluating GEC output, and we suggest that the GEC community adopt this modification 
to better assess systems.  

To facilitate GEC evaluation, we have set up an online platform\footnote[8]{\protect\url{https://competitions.codalab.org/competitions/12731}} for benchmarking system output on the same set of sentences evaluated using different metrics and made the code for calculating LT and LFM available.\footnote[9]{\protect\url{https://github.com/cnap/grammaticality-metrics}}


\section*{Acknowledgments}
We would like to thank Matt Post, Martin Chodorow, and the three
anonymous reviews for their comments and feedback.   We also
thank Educational Testing Service for providing \eratername output.
This material is based upon work partially supported by the NSF GRF under Grant No. 1232825.




\bibliographystyle{emnlp2016}
\bibliography{main}

\end{document}